\documentclass[letterpaper, 10 pt, conference]{ieeeconf}  

\IEEEoverridecommandlockouts                              
\usepackage{subfig}
\usepackage{subfloat}

\usepackage{amsmath,amsfonts,amssymb,amsthm,psfrag,enumerate}
\usepackage{graphicx}
\usepackage[framemethod=tikz]{mdframed}
\definecolor{mygray}{RGB}{248,248,250}
\newtheoremstyle{mystyle}
  {\topsep}
  {\topsep}
  {}
  {}
  {\bfseries}
  {.}
  {5pt plus 1pt minus 1pt}
  {{\color{black}\thmname{#1}~\thmnumber{#2}}\thmnote{\,--\,#3}}%
\theoremstyle{mystyle}
\newmdtheoremenv[%
  backgroundcolor=mygray,%
  linecolor=black,%
  leftmargin=0pt,%
  innerleftmargin=5pt,%
  innerrightmargin=5pt,%
  ]{probx}{Problem}

\title{\LARGE \bf
Predicting Injectable Medication Adherence via a Smart Sharps Bin and Machine Learning
}

\author{Yingqi Gu$^{1,\star}$, Akshay Zalkikar$^{2}$, Lara Kelly$^{2}$, Kieran Daly$^{2}$ and Tomas E. Ward$^{1}$
\thanks{$^{1}$Y. Gu, T. Ward are with the Insight Centre for Data Analytics, School of Computing, Dublin City University, Ireland.}%
\thanks{$^{2}$A. Zalkikar, L. Kelly and K. Daly are with HealthBeacon Ltd, Ireland.}%
\thanks{$^{\star}$Corresponding author. Email: {\tt yingqi.gu@dcu.ie}}%
\thanks{Y. Gu and A. Zalkikar are joint first authors in this paper.}
}

\begin{document}

\maketitle
\thispagestyle{empty}
\pagestyle{empty}


\begin{abstract}
	
Medication non-adherence is a widespread problem affecting over 50\% of people who have chronic illness and need chronic treatment \cite{world2003adherence}.  Non-adherence exacerbates health risks and drives significant increases in treatment costs. In order to address these challenges, the importance of predicting patients' adherence has been recognised. In other words, it is important to improve the efficiency of interventions of the current healthcare system by prioritizing resources to the patients who are most likely to be non-adherent. Our objective in this work is to make predictions regarding individual patients' behaviour in terms of taking their medication on time during their next scheduled medication opportunity.  We do this by leveraging a number of machine learning models. In particular, we demonstrate the use of a connected IoT device; a ``Smart Sharps Bin'', invented by HealthBeacon Ltd.; to monitor and track injection disposal of patients in their home environment. Using extensive data collected from these devices, five machine learning models, namely Extra Trees Classifier, Random Forest, XGBoost, Gradient Boosting and Multilayer Perception were trained and evaluated on a large dataset comprising 165,223 historic injection disposal records collected from 5,915 HealthBeacon units over the course of 3 years. The testing work was conducted on real-time data generated by the smart device over a time period after the model training was complete, i.e. true future data. The proposed machine learning approach demonstrated very good predictive performance exhibiting an Area Under the Receiver Operating Characteristic Curve (ROC AUC) of 0.86.\newline

\end{abstract}

\section{Introduction}

Medication non-adherence is a major concern worldwide. According to a report from the World Health Organisation (WHO) \cite{world2003adherence}, only 50\% of people adhere to their chronic therapy. Such poor medication adherence can affect the effectiveness of prescribed treatments, increasing safety risks for patients and creating unnecessary financial burden for the clinicians, healthcare industry and other stakeholders \cite{cutler2010thinking}. In this paper, adherence is defined as the degree to which the person's behaviour corresponds with the agreed recommendations from a health care provider \cite{world2003adherence}. In order to solve the problem of non-adherence, the first challenge to overcome is selecting an accurate method of measuring and monitoring patient adherence. A wide range of adherence measurement approaches have been reported in the literature \cite{anghel2019overview}, and these can be classified into two categories: the indirect method and the direct method. In particular, indirect methods include pill counts, self-reported questionnaires and Medication Event Monitoring System (MEMS). MEMS is considered the gold standard in terms of existing adherence measurement approaches and works through tracking the dates and time stamps when the patients open the bottle cap \cite{diaz2001use}. Compared to MEMS, adherence was roughly overestimated by 17\% using self-report and 8\% using pill counts \cite{tricco2016scoping}. The benefits of the self-report method are that it is both easy and straightforward for the patient. However, as the reported adherence can be biased by patients' inaccurate recall, thus the adherence is significantly overestimated \cite{nieuwenhuis2012self}. Even the gold standard MEMS approach faces some challenges, such as the potential for loss of data and limited user acceptance; some patients reported that they found it difficult to transfer medication into the MEMS bottle \cite{shellmer2007challenges}. Direct methods include measurement of drug levels in body fluids such as plasma and urine. The accuracy of this method may be influenced by drug metabolism so patterns of adherence cannot be obtained in this way. Moreover, direct observations can only be taken from hospitalized patients, so is not of relevance to those who take medications in a home setting.

Many medications for illnesses like arthritis, diabetes and multiple sclerosis are self-administered by injection at the patient's home: away from a supervised healthcare provider's environment. In order to better measure patient adherence and manage these self-administered injections, HealthBeacon have developed a connected IoT device, the ``Smart Sharps Bin'' (SSB). HealthBeacon is a digital medication technology company that develops smart tools for managing medication. The SSB is suitable for use in a home setting and digitally records the disposal of used hypodermic needles, passively tracking injection disposal data and uploading it to a cloud-based database for long-term storage and further analysis to enable the support team to monitor the drop disposals and provide follow-up services as necessary. The use of the SSB doesn't require any major alteration to standard hypodermic needle disposal behaviour, and brings additional benefits to the patient: the system prompts the patient when their next injection is due and reports their medication adherence status, without the need for direct interaction with the healthcare provider. The SSB has FDA clearance and has been integrated into patient care programs throughout North America and Europe. It can therefore be considered as state of the art technology for measuring patients' adherence accurately and efficiently. In order to improve the patient adherence rate, precise measurement of medication adherence and the use of appropriate intervention methods (by the healthcare system) are important. Another key factor in this context is about predicting the accuracy of patient adherence in the future. If the patients that are most likely to be non-adherent can be accurately identified and targeted, the efficiency of interventions can be significantly increased \cite{cutrona2012targeting,nelson2019predicting}. Non-adherence is a multifactorial problem that can be influenced by a range of patient-, disease-, condition-, social/economic- and healthcare system-related factors \cite{cea2019association}. It is common for patients to be confused by treatment schedules, forget to take their medications due to unexpected events, discontinue to take their medication due to side effects, or stop taking the treatment because they feel they no longer need it. For predicting patient adherence, significant effort has been devoted to investigation of the correlation or relationship between these various factors and the level of adherence \cite{steiner2009sociodemographic,osterberg2005adherence}. In the work by Schuz \textit{et al.} \cite{schuz2011medication}, the authors attempted to predict patient adherence by analysing medication beliefs, showing that a patient's beliefs about medication affect both intentional and unintentional treatment adherence. Stilley \textit{et al.} \cite{stilley2004psychological} predicted patient adherence based on patient-related features such as gender, age and race. While many of these features correlate with patient adherence, they demonstrate weak discrimination between adherers and non-adherers. In contrast, prediction of adherence based on prior pill refill data achieved higher accuracy \cite{molfenter2012roles}, illustrating that historic adherence records contain more accurate predictive information of future patient treatment adherence. More recent work has applied emerging machine learning techniques rather than analysis of patient characteristics to predict patient adherence. In a small scale study of Parkinson's disease patients, Tucker \textit{et al.} \cite{tucker2015machine} predicted patient adherence by adopting a remote data mining approach to analyse whole-body movement data collected by a non-wearable hardware device, and subsequently classifying the patient using machine learning methods. Karanasiou \textit{et al.} \cite{karanasiou2016predicting} applied eleven classification algorithms such as SVM and Bayesian Networks to predict the adherence of patients with heart failure based on a dataset of 90 patients, and Franklin \textit{et al.} \cite{franklin2016observing} aimed to predict patient adherence in the next 30, 60 and 90 days using ten different machine learning models based on information from Medicare enrolment files and medical and pharmacy claims. 

In this work, historic hypodermic disposal data collected from patients' SSBs has been selected as the primary variable for adherence prediction; analysis of the importance of various different data features (features generated by HealthBeacon's SSB and management system) in terms of future adherence prediction is discussed in detail below. Using a large dataset with 5,915 HealthBeacon units and 165,223 historic hypodermic drop records over a three year period, a patient adherence prediction model was developed. To the best of our knowledge, this work is the first machine learning-based prediction model to date to be developed using an extensive, real patient dataset. Specifically, our work has 3 main contributions:

\begin{itemize}
	\item[(i)] We proposed and developed the SSB, a connected IoT device, which can be used to record a patient's historic hypodermic drops (referred to as a `drop' from hereon in, drop is defined as the event of disposing the hypodermic medication into SSB) and monitor patient adherence in their own home. The raw dataset collected through the SSB contains a substantial amount of valuable data with various features including drop date/timestamp.\newline

    \item[(ii)] We trained and developed machine learning models in an end-to-end fashion using the data collected by the SSB. We propose an ensemble learning system that combines 5 different machine learning models to generate an adherence prediction model capable of predicting a patient's adherence at their next scheduled medication event. \newline

    \item[(iii)] We performed experiments and testing on real world self-medicating patient data from a unseen dataset and demonstrated that our approach achieved both excellent prediction performance and good generalisation with an accuracy of 81.24\% and an AUC of 0.86, respectively. 
\end{itemize}

Technical details of the ``Smart Sharps Bin'' are given in Section \ref{SB}, the methodology of machine learning model construction is presented in Section \ref{Method}, and the testing work and results are discussed in Section \ref{results}. General conclusions can be found in Section \ref{Conclusion}. \newline

\section{Data Collection: the Smart Sharps Bin (SSB)} \label{SB}

The SSB is an injectable medication management system which can be easily integrated into a patient's home and constantly monitor the medication disposal. For further clinical intervention, we predicted the patient medication adherence by leveraging machine learning models based on the real time data collected from the SSB. The device was launched in May 2015 by HealthBeacon Ltd., and since then it has been used by 9,000+ patients and has tracked 300,000+ disposed hypodermic needles. \newline

\subsection{SSB Technical Details}

\begin{figure}[htbp]
	\begin{center}
		\includegraphics[width=3.5in, height = 2.1in]{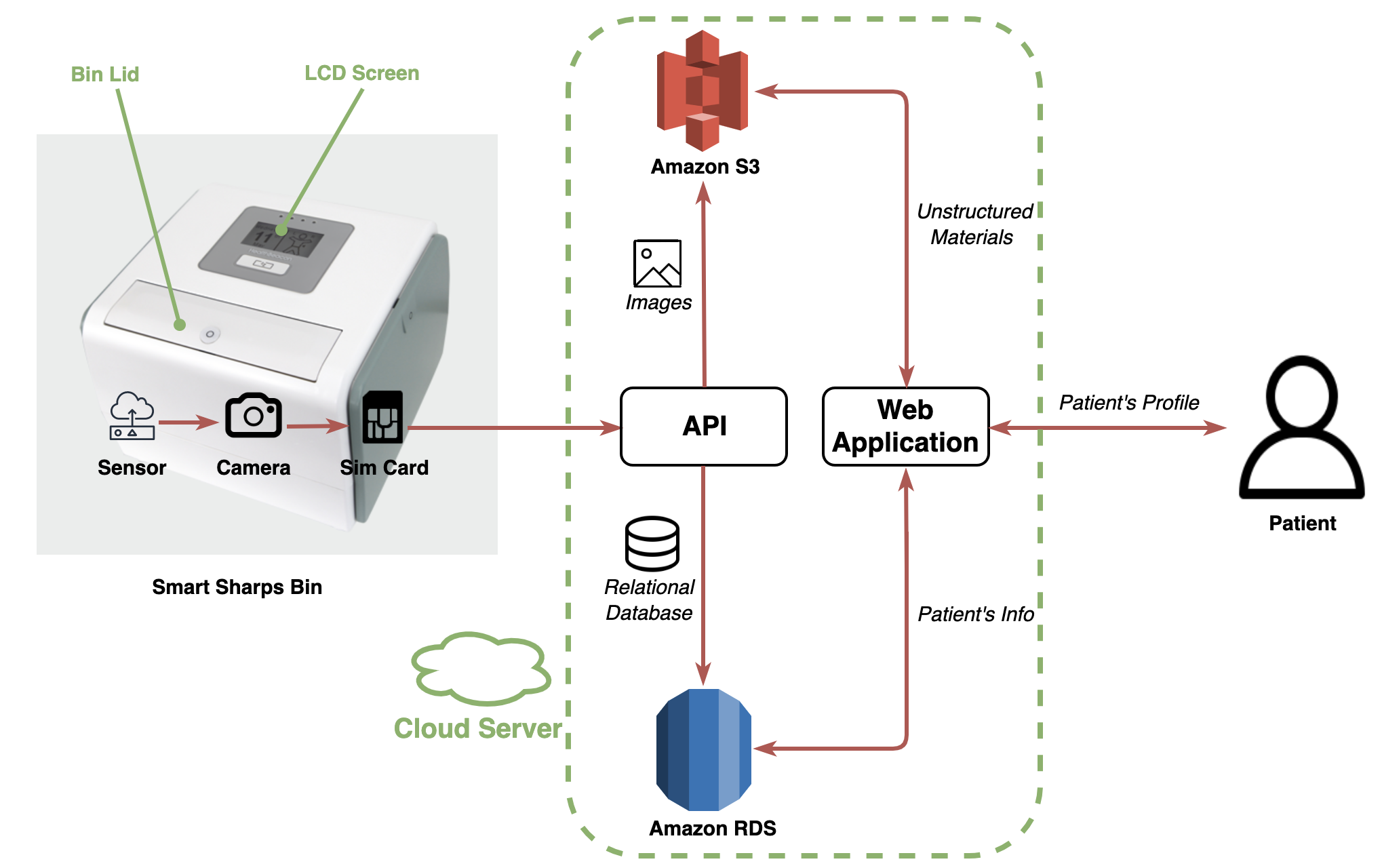}
		\caption{A schematic diagram of the HealthBeacon SSB.}\label{smartbin}
	\end{center}
\end{figure}

The SSB has been designed as a cuboid and consists of five main parts, as shown in the schematic diagram in Figure \ref{smartbin}. The uppermost surface contains an LCD screen and an integrated sharps bin lid. When a patient starts a treatment programme suitable for at-home medication through injection, an SSB is dispatched to the patient, pre-programmed with their personal medication schedule information, including information such as the scheduled start date, injection frequency, required injection location (area of body) and preferred short message service (SMS) time slots. When an injection is due, the blue light notification above the LCD screen lights up as a visual reminder for the patient. Along with this, a reminder SMS is sent to the patient at their preferred time slot. If the patient fails to take their medication/forgets to drop it into the HealthBeacon SSB, an intervention SMS is issued the following day. The SMS' reminder system works in a smart way as it only triggers SMS when required. The left zone of the LCD screen shows the adherence score of the patient: initialized as 0\% at the very beginning of the treatment, the score increases or decreases according to patient's adherence rate over a given period, to encourage the patient to stay on track with their medication. When the patient is ready to inject, they administer the medication at the injection location shown on the LCD screen. After injection, the patient disposes of the used injector by pushing it through the bin lid (a traditional hazardous sharps bin sits inside the SSB). When an injection is dropped into the SSB, a sensor beam which is fitted inside the lid is tripped and triggers a micro-camera inside the SSB to capture a picture of the injection along with a time stamp. For each drop, 2 files are created: a .csv file containing the time stamp and a .bmp file with the image. The files are then uploaded to HealthBeacon's cloud server through a secure private Access Point Name (APN) via a Machine to Machine (M2M) simcard. All data is encrypted both in transit and rest based on Advanced Encryption Standard (AES) 256 . The unstructured dataset is finally uploaded to Amazon Simple Storage Service (S3) for long-term storage while the dataset saved as CSV files is migrated to Amazon Relational Database Service (RDS). Complimentary to the SSB data, a web application has been developed where patients' limited personal data including age, name, medication type and injection frequency can be collected for programming the SSB set-up for each patient.

\section{Methodology} \label{Method}

\subsection{Data Extraction and Preprocessing}
HealthBeacon's cloud-based database includes information regarding patients and their trearment programme, from the scheduled medication date and frequency for each patient to the drop status of scheduled injections. The dataset used for training the predictive machine learning models was collected from the 7th May 2015 to 27th Oct 2018, inclusive. The dataset contains information obtained from 5,915 HealthBeacon units, including 165,223 individual drop events. The data was retrieved and exported from the database in the format of CSV files for further processing.

Only the essential anonymised data fields from the web application were extracted in order to comply with EU General Data Protection Regulation (GDPR) policies \cite{gdpr}. The dataset was then cleaned to remove empty rows/columns and redundant variables. Categorical data other than drop status was transformed into a machine learning-compatible format using one hot encoding. One hot encoding is a classic technique applied in machine learning that converts categorical variables into a form that is suitable for provision to machine learning algorithms. 

The time stamp of the drop status has been converted to labels as ``On-Time'' / ``Not On-time'' based on the time difference between the drop time stamp and the scheduled medication time stamp.

\begin{figure}[htbp]
	\begin{center}
		\includegraphics[width=2.2in, height = 1.2in]{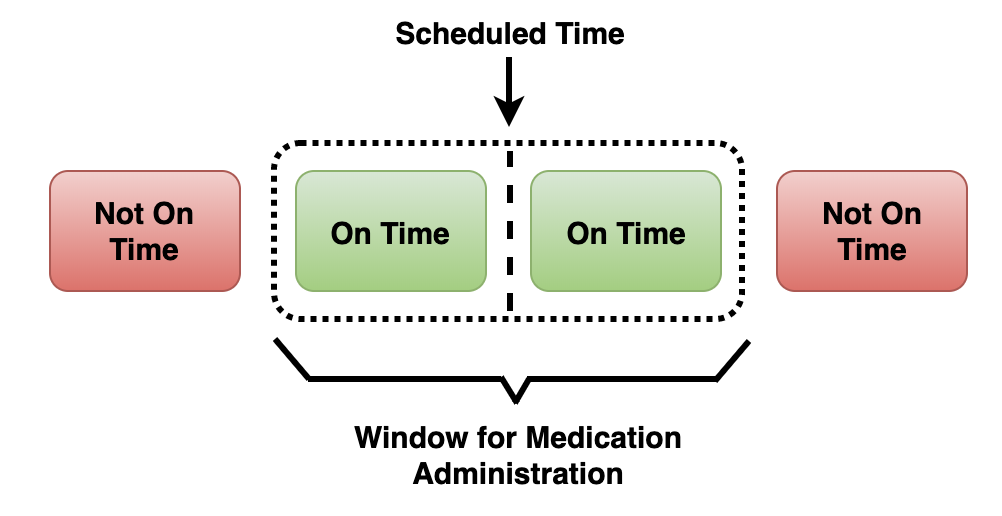}
		\caption{The time frame defined for ``On-Time'' and ``Not-On-time'' disposals.}\label{time}
	\end{center}
\end{figure}

We define the recommended time period for medication by the medical prescription as  "Window for Medication Administration (WMA)" in this paper. WMA varies by different drugs and different scheduled medication frequency. In the context of this work, the drop was labelled as ``On-Time'' if the medication was taken within the window in SSB, otherwise, the drop was marked as ``Not On-time'' if the medication was taken outside the window or never dropped in the SSB at all. Thus, the problem has been formatted as a binary classification problem. WMA and the drop status labels are shown in Figure \ref{time}.

\subsection{Feature Selection}

The next step was to decide which features should be taken into account for this problem and then feed them into machine learning models. Firstly, in order to select important features, we used Waikato Environment for Knowledge Analysis (WEKA) \cite{holmes1994weka}, an open-source software containing a collection of visualization tools and algorithms for data analysis and machine learning modelling. WEKA helped to filter out any uncorrelated features, only selecting important features that contain the best predictive information using the information gain-based feature selection technique and the built-in 'InfoGainAttibuteEval' attribute evaluator. After running the evaluator on the entire training dataset, the selected features can be mainly classified as historic drops, medication frequency and country.

\begin{figure}[htbp]
	\begin{center}
		\includegraphics[width=3.1in, height = 1.85in]{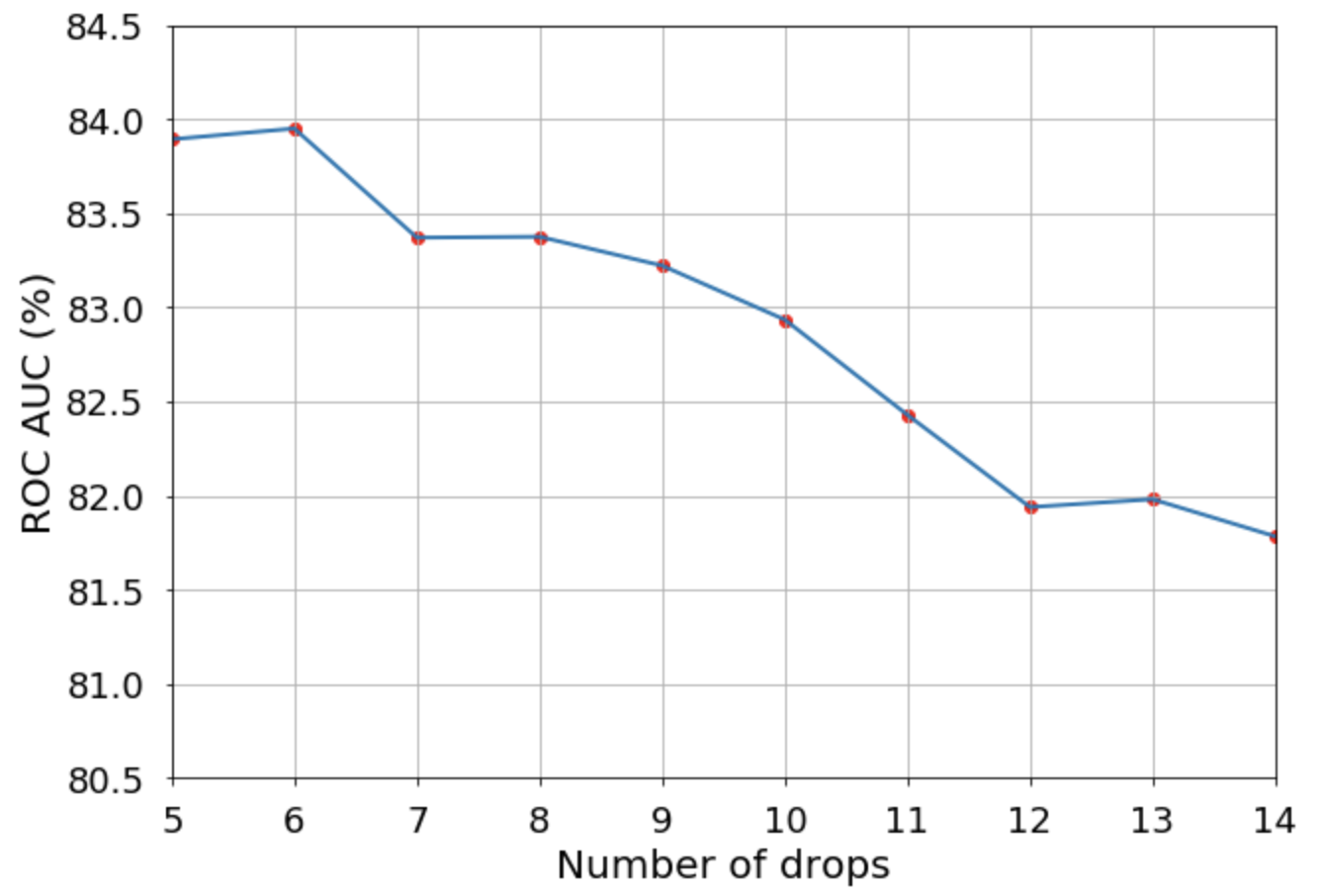}
		\caption{The ROC AUC score versus the number of drops. The ROC AUC was computed based on the Random Forest model by incrementally selecting a number of historic drops ranging from 5 to 14.}\label{drops_roc}
	\end{center}
\end{figure}

It was observed that patient drop history is the most significant feature, with the most recent drop being the most important. Thus, historic drop data was taken as a very significant attribute for prediction of medication adherence. In order to decide the number of historic drops to take into account as features, we incrementally increased the number of historic drops from 5 to 14 and attempted to predict the status of the next drop. Moreover, we calculated the ROC AUC score, varying the number of drops from 5 to 14 evaluated on Random Forest algorithm. The performance of the Random Forest model was evaluated by progressively increasing the number of drops from the last (most recent) five drops up to the last fourteen drops. The ROC AUC score was plotted in Figure \ref{drops_roc}, and shows that the ROC AUC score is largest when six historic drops are taken into account. Thus, we chose to use the last 6 historic drops in order to predict patient adherence for the next scheduled medication event. Therefore, the units with less than 7 historic drops attached to their record were removed from the dataset. This reduced the number of samples in the training set to 160,865 (drops) associated with 4,609 units. 

\begin{figure}[htbp]
	\begin{center}
		\includegraphics[width=3.2in, height = 1.85in]{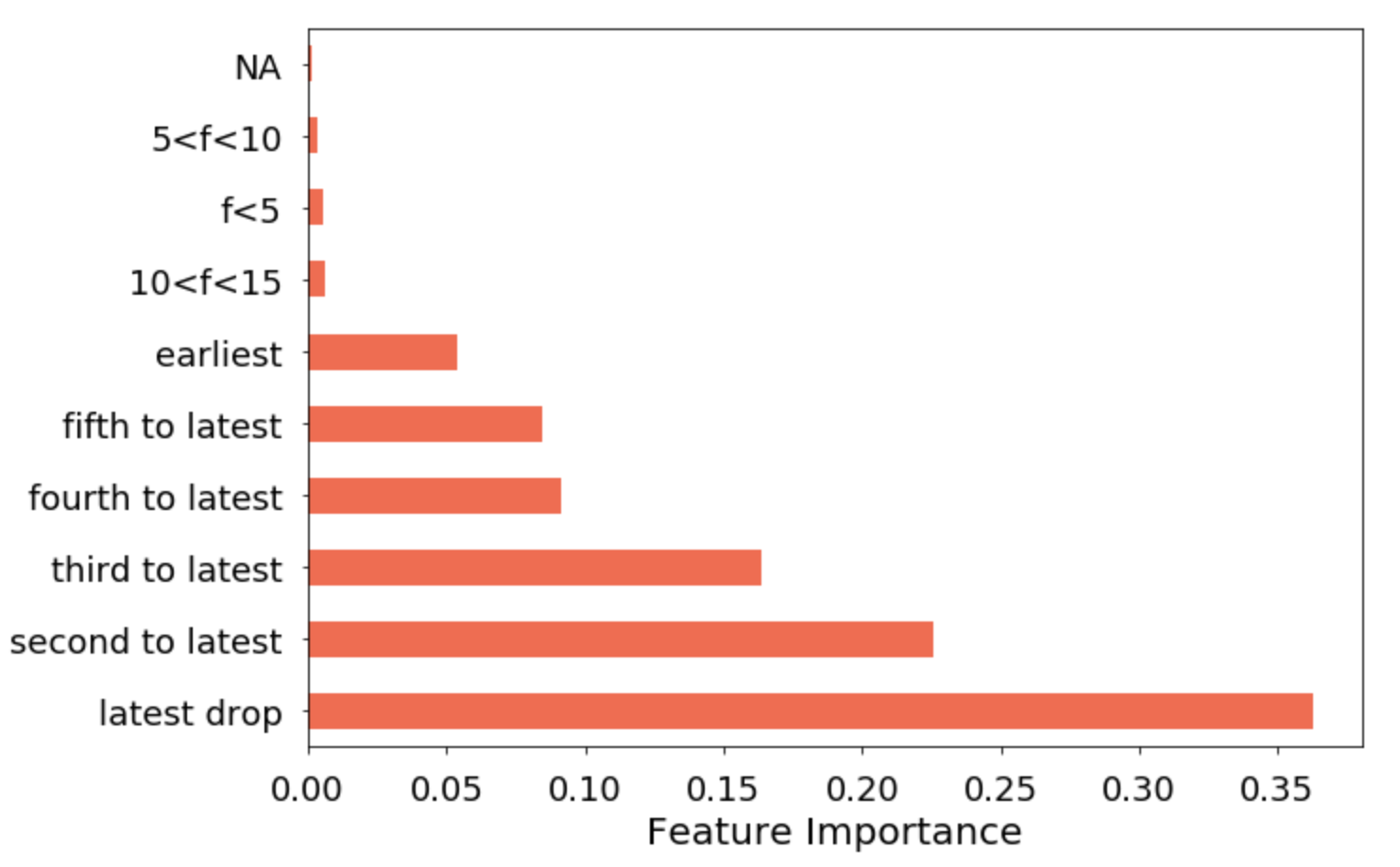}
		\caption{The top 10 most important features, selected using Extra Trees Classifier. Note that ``latest drop'' in this figure refers to the drop the patient disposed in last scheduled date while ``earliest'' means the first drop made in the sequence of 6 historic drops. ``f'' stands for frequency and ``NA'' is the abbreviation of North America.}\label{features}
	\end{center}
\end{figure}

\begin{figure}[htbp]
	\begin{center}
		\includegraphics[width=3.3in, height = 1.9in]{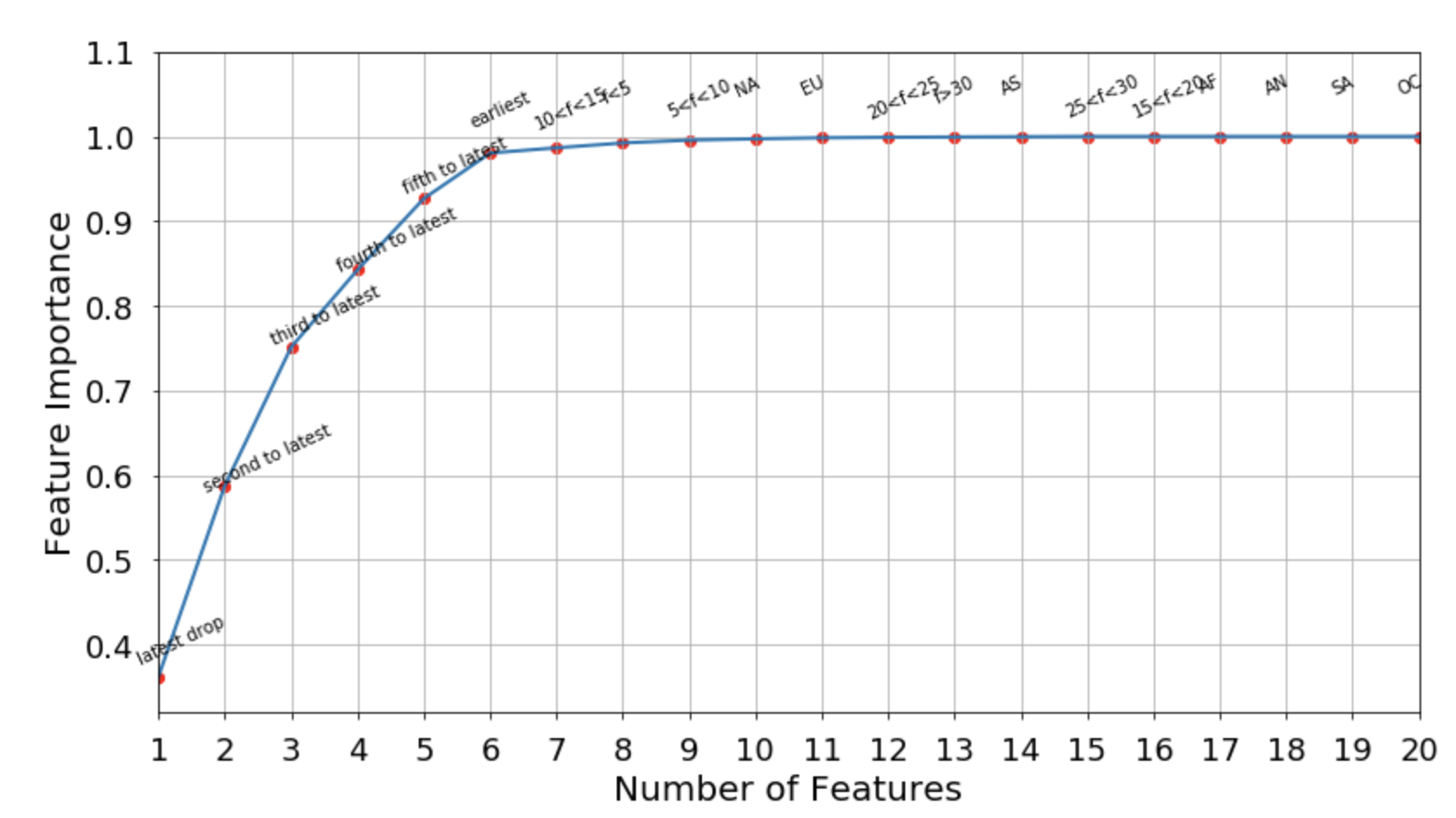}
		\caption{The accumulated importance of all features (NA: North America, EU: European Union, AS: Asia, AF: Africa, AN: Antarctica, SA: South America, OC: Oceania).}\label{accumulated features}
	\end{center}
\end{figure}

Furthermore, the top 10 most important features have been plotted by using the Extra Trees Classier algorithm in Figure \ref{features}, and the accumulated importance of features is also demonstrated in Figure \ref{accumulated features}. From these two Figures, it is clear that historic drops are the dominant features, constituting 99\% of the feature importance, meaning that historic drop data contains the most useful predictive information, while the second most important feature is medication frequency. Here frequency at which patient is scheduled to administer their injection at a regular basis, for example, once a week or daily.

\subsection{Modelling and Hyper-parameters Tuning}

Firstly, we split the entire dataset into two subsets: training set and validation set. In our project, the training data set was used to fit a group of candidate machine learning models and the validation set was used to optimize the models by tuning parameters and hyper-parameters. The total dataset was split and allocated randomly between these two categories with the following proportion: 80\% for training set and 20\% for validation set.

\begin{figure}[htbp]
	\begin{center}
		\includegraphics[width=3.65in, height = 1.7in]{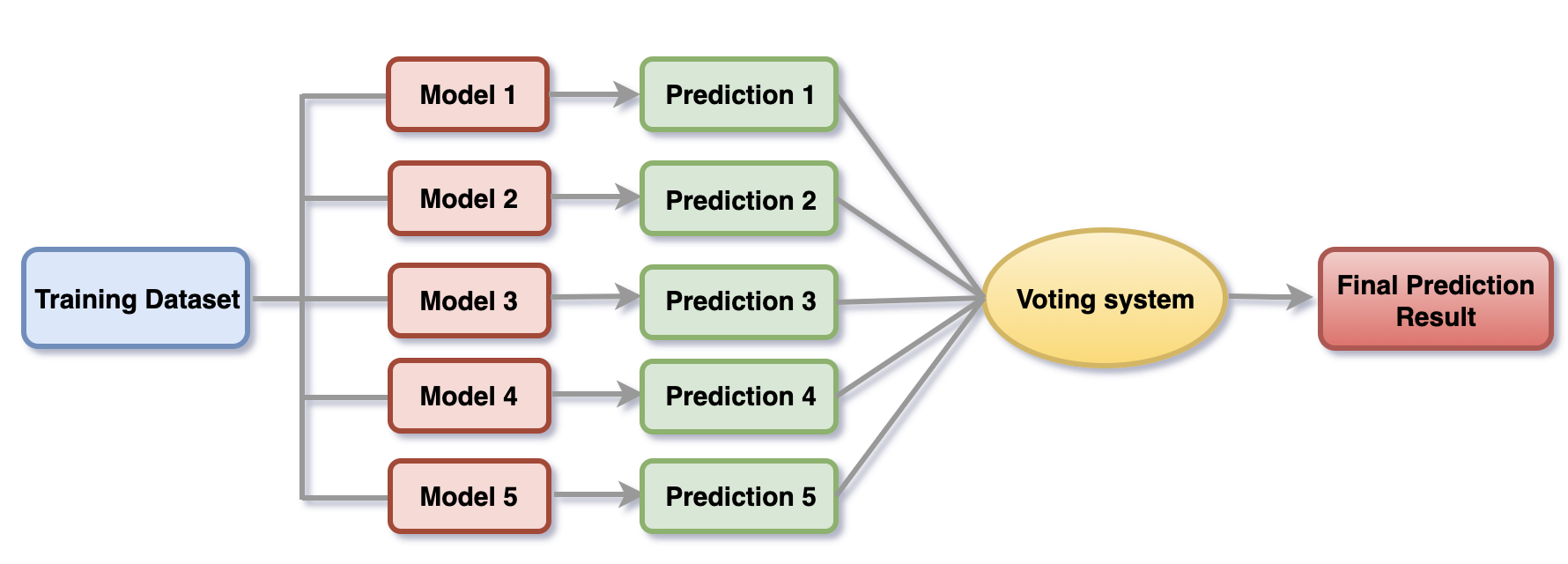}
		\caption{The ensemble learning and majority voting system. We selected the majority of the binary prediction results as final result through the voting architecture.}\label{voting}
	\end{center}
\end{figure}

The main objective of this project was to predict a binary classification of whether a patient is going to take medication ``On-Time'' (within the WMA) or ``Not On-Time'' (outside the WMA). To increase the prediction accuracy and to avoid the risk of overfitting, we selected, constructed and evaluated several standard ensemble learning models and artificial neural networks. Ensemble learning combines multiple machine learning models for better predictive performance and decreased likelihood of overfitting (lower variance), as different models generally do not make all of the same errors on the testing set \cite{dietterich2002ensemble}. The hyper-parameters for the models were optimised by applying both Random Search and Grid Search with the 10-fold cross validation method. After evaluation using the validation dataset, we selected the top five performing models: Extra Trees Classifier, Random Forest, XGBoost, Gradient Boosting and Multilayer perception. Furthermore, we combined these five models and proposed a majority voting architecture where the majority of the prediction results obtained from the five models will be adopted as the final result. For instance, if the prediction results from the five models are displayed as: ``On-Time'', ``Not On-Time'', ``Not On-Time'', ``On-Time'' and ``On-Time'', the final prediction result would be ``On-Time'' as the result included three occurances of ``On-Time'' compared to two counts of ``Not On-Time''. A schematic diagram illustrating this process is shown in Figure \ref{voting}. We used Scikit-learn \cite{pedregosa2011scikit}, a free open-source machine learning library for Python, to construct the machine learning models and search for optimal parameters and hyper-parameters. \newline

\subsection{Model Evaluation and Testing}
Drop data was recorded by the SSB and uploaded to the database for storage on a daily basis. This daily-generated dataset was retrieved for the period from 20th December 2018 to 7th March 2019 and used as the testing dataset, as using data previously unseen by the models means that the testing work is free from `data leak' problems. During this period, for patients expected to take medication, the `Prediction file' containing the historic drop information of the six most recent drops was collected at the beginning of the scheduled medication day. This data was used to predict if the 7th (next) drop would be ``On-time'' or ``Not On-time''. Once the `Prediction file' had been extracted, it was fed to the proposed machine learning models in order to predict which category of classification (``On-Time'' / ``Not On-Time') would be allocated to the patient that day.

As we mentioned above, WMA is the recommended time range within which a drug should be administered. The WMA can vary for different drugs and also for different scheduled medication frequencies. For one drug type involved in the study the adherence window is 144 hours, which means all drops made within 72 hours of the scheduled medication time were labelled as ``On-Time'', in this case, in order to give patients enough time to take medication and dispose of the used injector, the `Testing file' (including the ground truth on the real behaviours of patients) was collected 72 hours after the scheduled medication day. The testing work compared the `Prediction file' with the `Testing file'. For simplicity, we assume an example where a patient is scheduled to take medication on the 5th January 2019; the `Prediction file' is generated at 12 am on the 5th January 2019 and contains data from the last 6 drops; if the window is 144 hours, the `Testing file' with the real-world drop result is then collected at 12 am on 8th January 2019. During the period from 12 am on 2nd January to 12 am on 8th January, if the patient takes the medication and makes the right disposal within the 144 hours, the behaviour can be labelled as ``On-Time'', otherwise, the drop is classified as ``Not On-Time''. This data was recorded in `Testing file' as ground truth for further testing work.

Before conducting the testing work, we removed data generated in the following scenarios:

\begin{itemize}
	\item Unplugged: if the patient's unit is unplugged for a period of more than 30 days before the scheduled medication drop day. Please note that in principle the SSB is required to remain plugged in all the time for constant monitoring. However, we consider it as unplugged when the unit is not communicating for a period of more than 30 days.

    \item Deactivated: if the patient's unit is deactivated in the period between extraction of the `Prediction file' and `Testing file' (excluded from the dataset due to insufficient knowledge regarding the ground truth of the patient's test drop).

    \item Self-Reported: In an event of the patient being away from the SSB, their drop status can be reported independently after the scheduled drop time. Inclusion of self-reported drops/amended drops are not suitable as there is too much uncertainty in patient behaviour and only the accurate drop information is considered as the ground truth for testing.

    \item Loading Dose: a loading dose is an initial higher dose of medication or series of such doses given in order to rapidly achieve a therapeutic concentration in the body. Loading doses were excluded as it would introduce a bias in predicting the future drops. \newline

\end{itemize}

\section{Results} \label{results}

\begin{figure}[htbp]
	\begin{center}
		\includegraphics[width=3in, height = 2.05in]{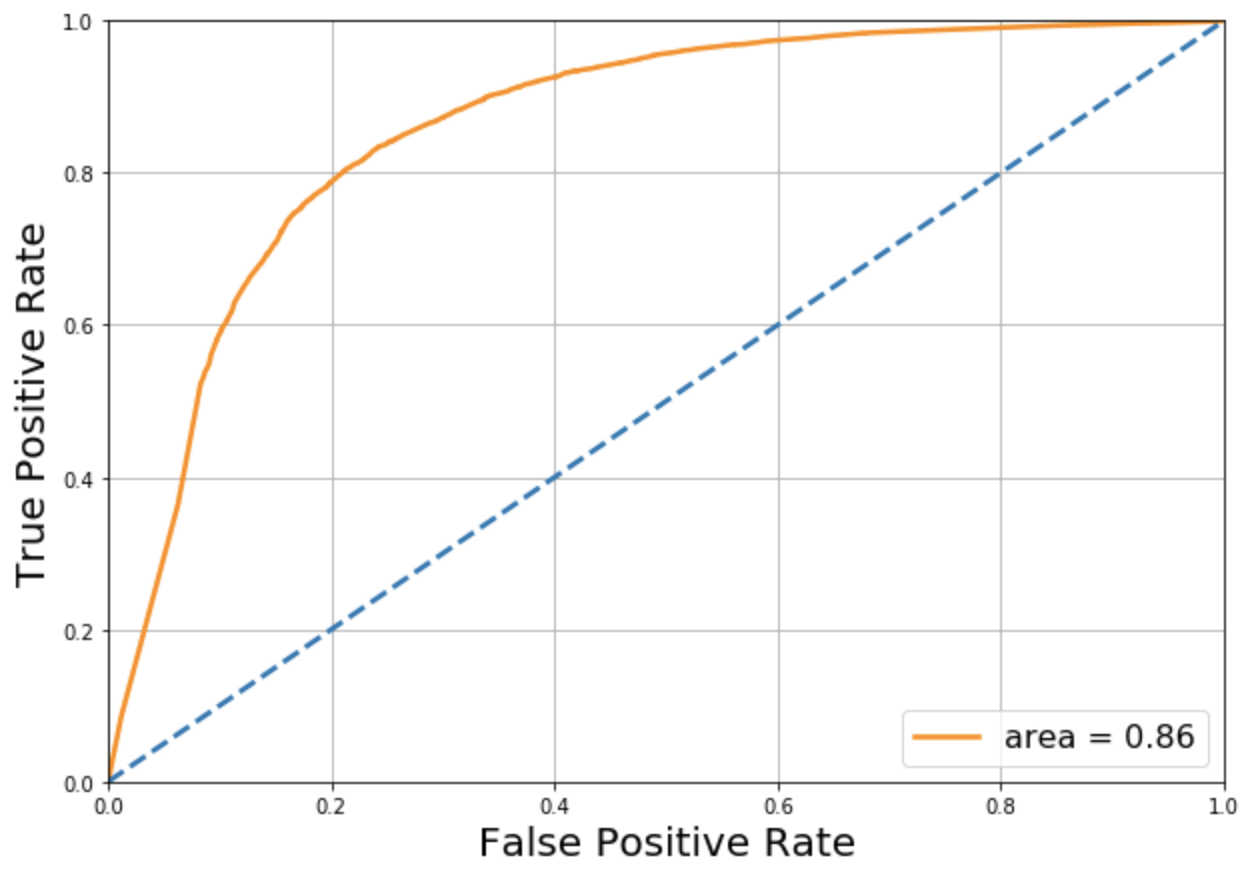}
		\caption{ROC curve for performance on the testing set (AUC = 0.86).}\label{ROC}
	\end{center}
\end{figure}

\begin{figure}[htbp]
	\begin{center}
		\includegraphics[width=3.4in, height = 2.3in]{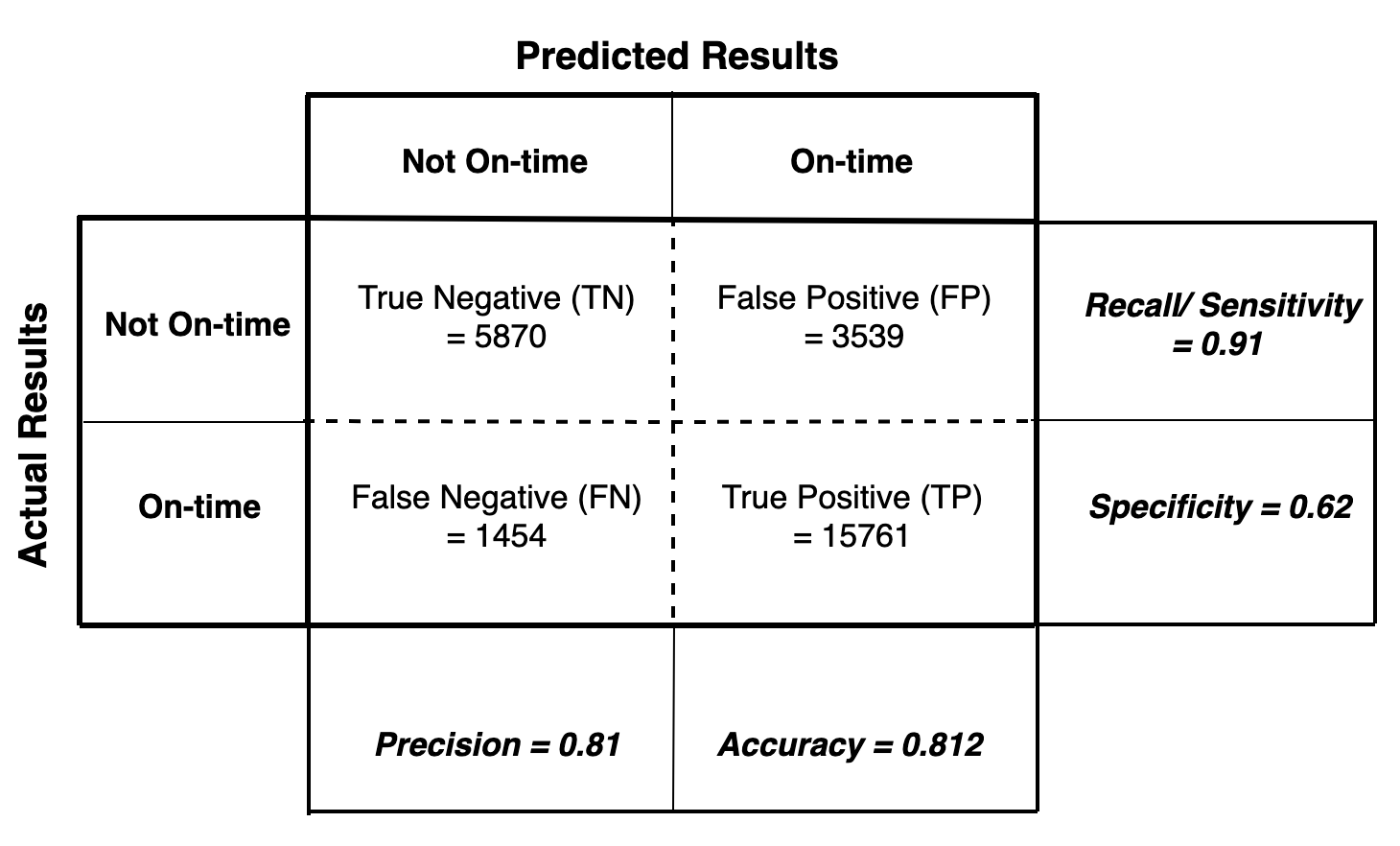}
		\caption{The confusion matrix generated for the testing set, with evaluated metrics.}\label{confusion}
	\end{center}
\end{figure}

\begin{figure}[htbp]
	\begin{center}
		\includegraphics[width=3.4in, height = 2.3in]{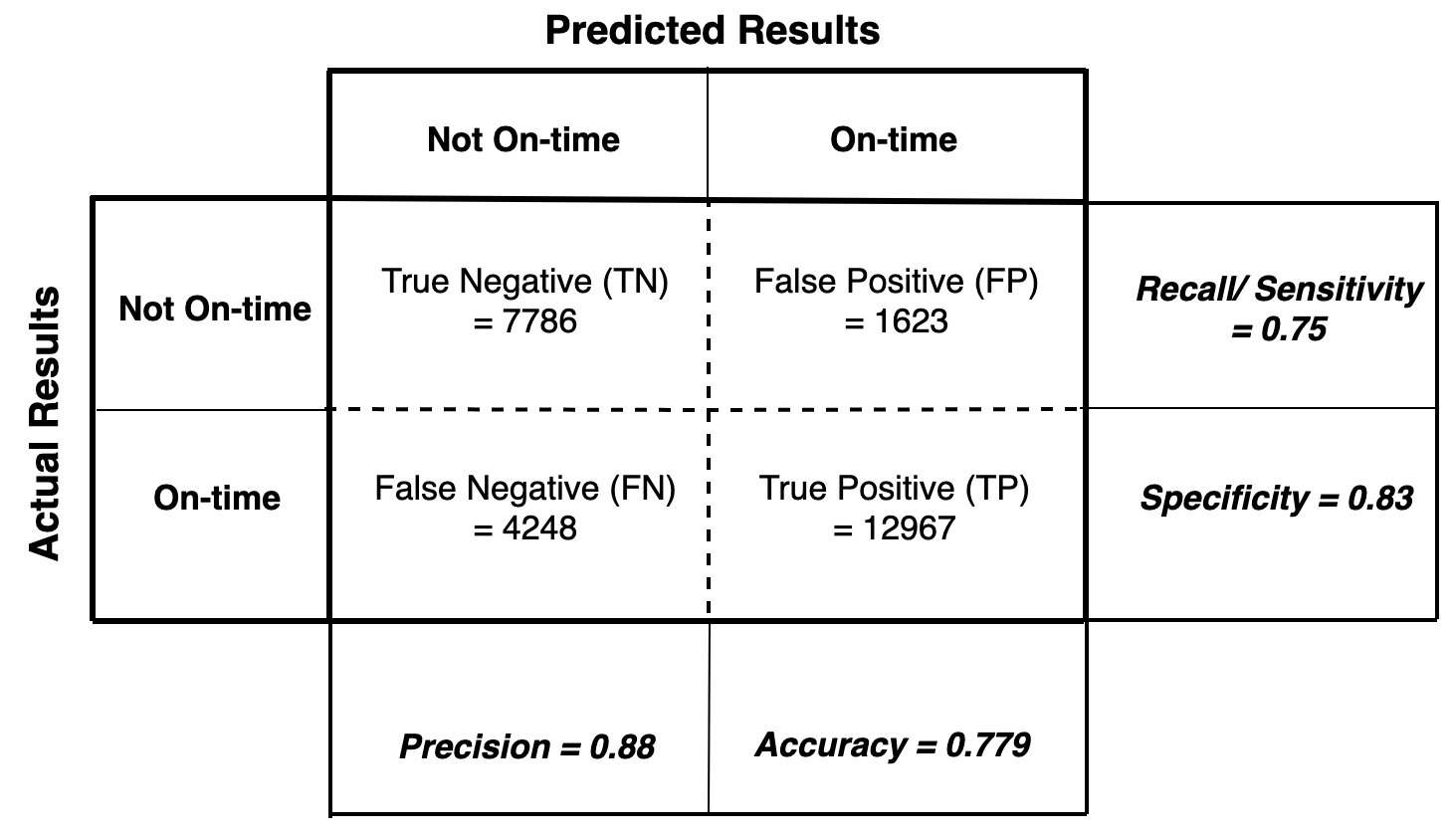}
		\caption{The confusion matrix generated for the testing set after resampling.}\label{confusion2}
	\end{center}
\end{figure}

In testing work, we evaluated the machine learning models by generating both a  ROC curve and a confusion matrix for better illustration. The ROC curve in Figure \ref{ROC} shows the AUC score of the predictive model to be 0.86 on the testing set. From the confusion matrix (the confusion matrix and additional performance metrics are illustrated in Figure \ref{confusion}) we can see 26,624 drops after exclusions were used for prediction. Among these, 21,631 drops were predicted correctly by using our trained models: a prediction accuracy of 81.24\% on the testing set. In addition, the recall/sensitivity from the confusion matrix is 91\%, which shows 91\% of the prediction made on those who will take medication on time (``On-Time'' type) is correct based on the constructed models. While the specificity exhibits the ratio of the correct prediction made on the people who will not take medication on time (``Not On-Time'' type) is 62\%. In the real world, our objective is predicting which patients are likely to not adhere to the medication so that interventions can be introduced to this group before the scheduled medication day in an attempt to improve the adherence rate. In other words, the prediction accuracy of these who are ``Not On-Time'' is more important in the case. Consider the number of samples in the ``On-Time'' class in our training set exceeded the ``Not On-Time'' class: the class imbalance can bias models to the majority class. In order to neutralize this, we retrained the models by randomly under-sampling the majority class. The confusion matrix generated from the retrained model in Figure \ref{confusion2} shows the specificity increases from 62\% to 83\% after resampling, which means that 83\% of patients who didn't take medication on time were predicted correctly according to the retrained model. 

\begin{figure}[ht]
	\centering
	\subfloat[Model performance based on `Accuracy'.]{\includegraphics[width=3in, height = 1.9in]{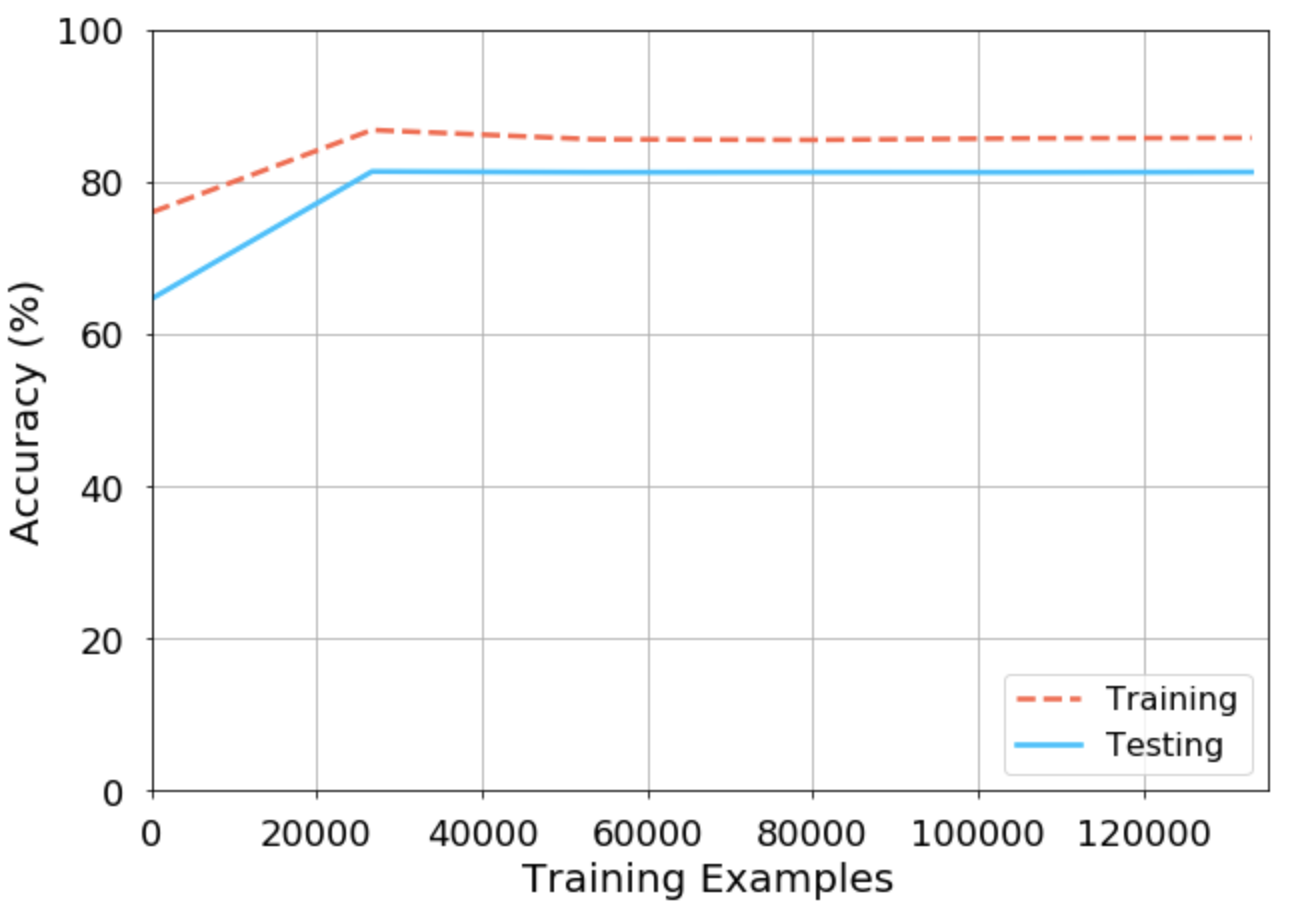}
		\label{figa}}
	\hfil
	\\
	\subfloat[Model performance based on `Precision'.]{\includegraphics[width=3in, height = 1.9in]{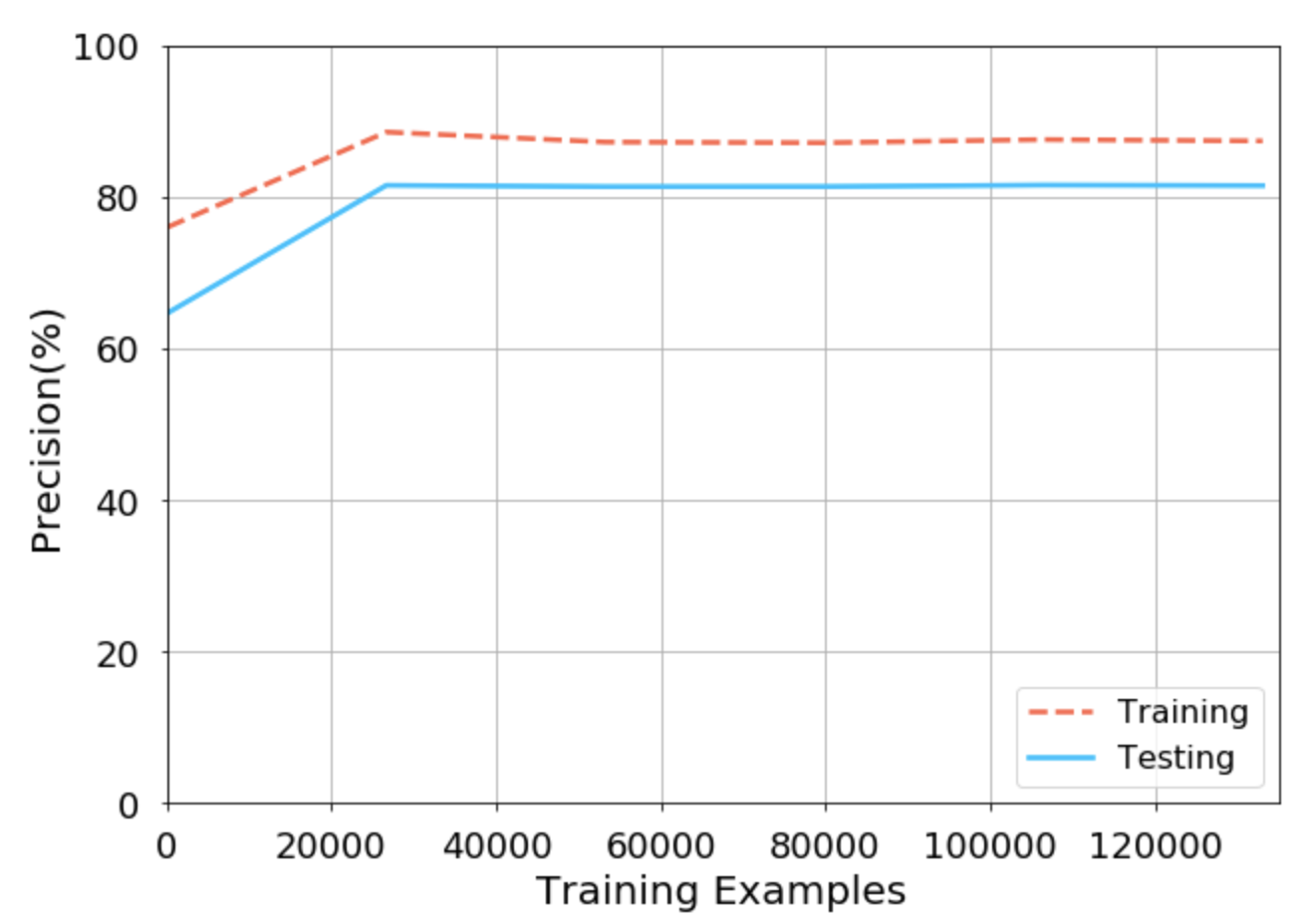}
		\label{figb}}
	\hfil
	\\
	\subfloat[Model performance based on `F1 score'.]{\includegraphics[width=3in, height = 1.9in]{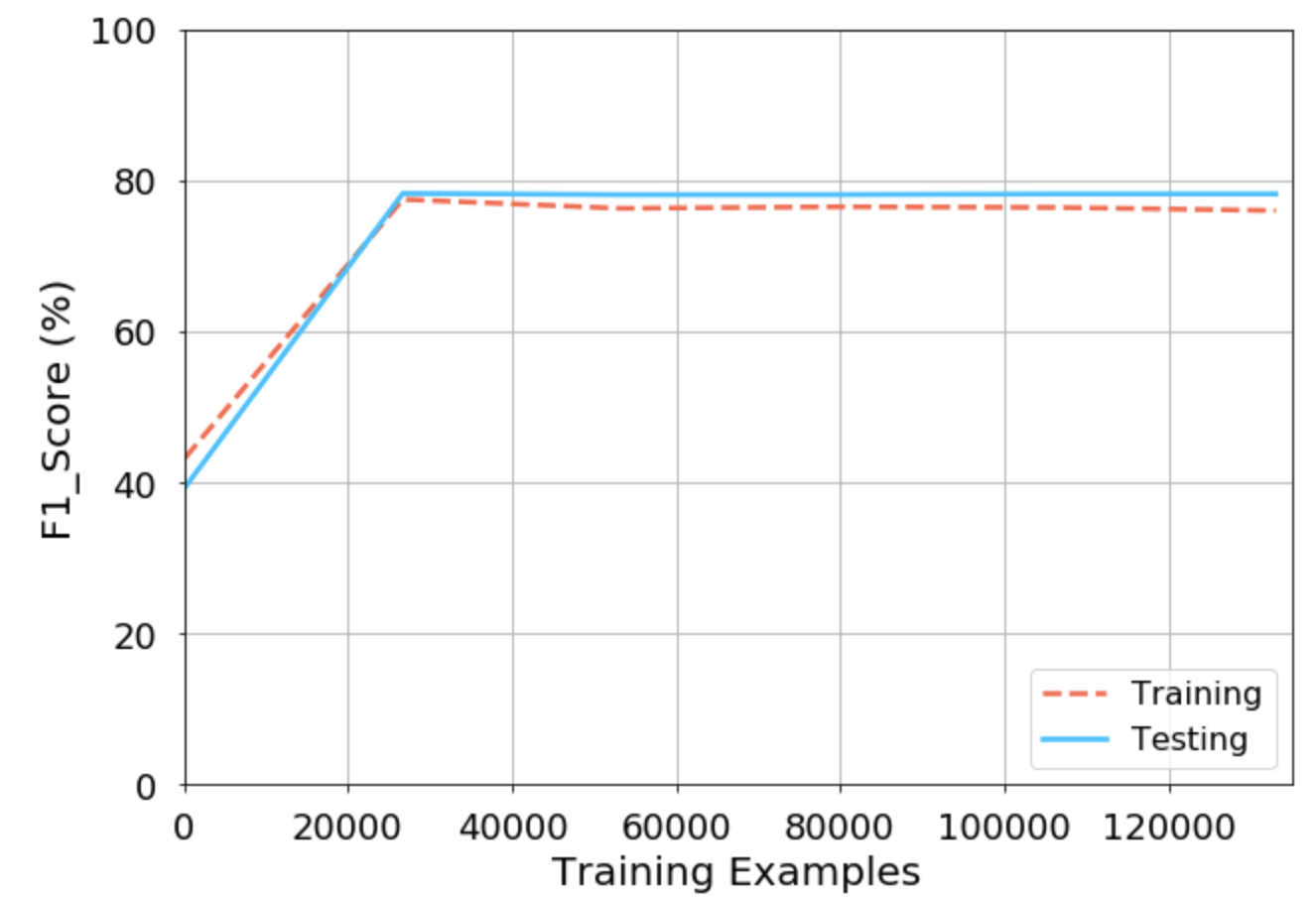}
		\label{figc}}
	\caption{Learning curves on both the training and testing dataset. The \textit{x}-axis represents the number of training samples and the \textit{y}-axis shows the performance of machine learning model.}
	\label{LC}
\end{figure}

\begin{figure}[htbp]
	\begin{center}
		\includegraphics[width=3.6in, height = 2in]{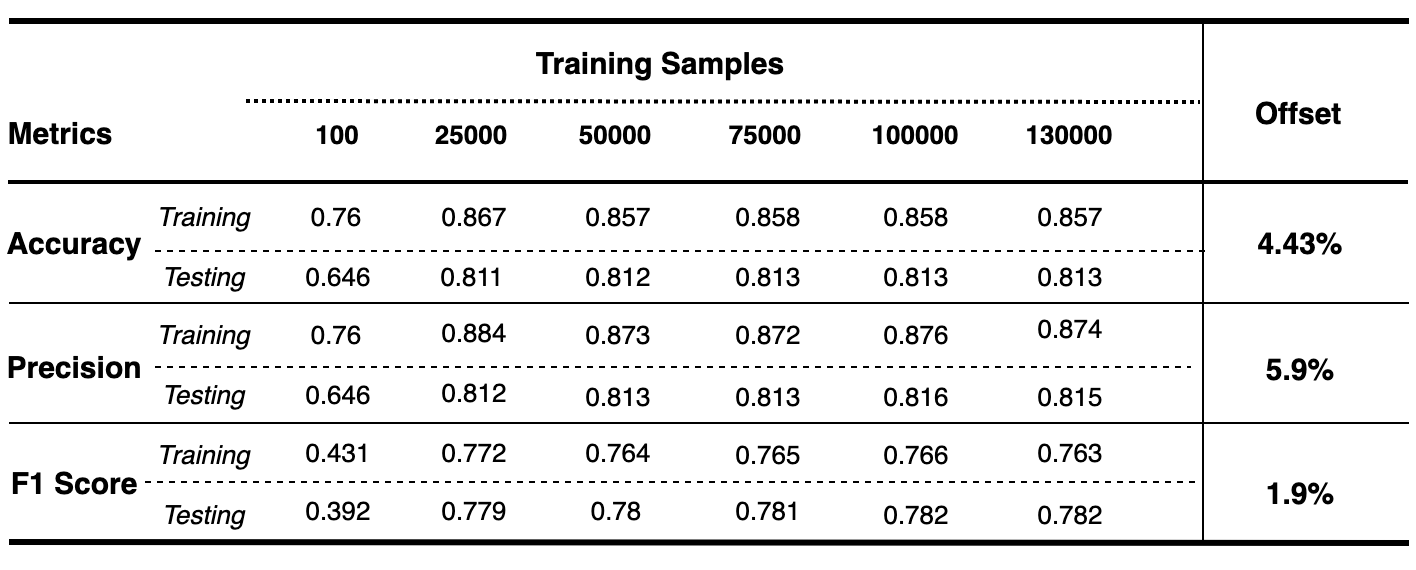}
		\caption{The accuracy, precision and F1 score on both training set and testing set at selected specific iterations: 100, 25,000, 50,000, 75,000, 100,000 and 130,000.}\label{Table}
	\end{center}
\end{figure}

In order to test if the constructed model is overfitting or underfitting, we generated `Learning Curves' as a diagnostic tool by plotting and comparing the performance of the proposed model on both the training set and testing set. `Learning Curves' are plots that show changes in learning performance over time in terms of experience \cite{lc}. By incrementally adding training samples, the model performance can be evaluated on both the training dataset and a hold out testing set after each update during the training process. In this work, we progressively increased the size of the training dataset from 100 training samples to 130,000 training samples for constructing the model, and then plotted the model performance on the training set and testing set versus dynamic training size. The accuracy, precision and F1 score have been selected as performance metrics and each metric has been plotted in Figure \ref{LC}. The `Train Learning Curve' demonstrates how well the model is learning and the `Test Learning Curve' gives an indication of the generalization performance of the model on unseen data. Based on this, an ideal result would be for the offset between these two lines to be as small as possible. We can see from the Figures that the offset between the two plots is very small. The concrete performance metrics are listed in more detail in Figure \ref{Table}. These values show the accuracy, precision and F1 score on both the training set and testing set at selected specific iterations: 100, 25,000, 50,000, 75,000, 100,00 and, 130,000. Figure \ref{Table} also shows that the offset between the training and testing dataset based on accuracy, precision and F1 score are 4.43\%, 5.9\% and 1.9\% respectively, which illustrates the constructed model is in general a good fit with suitable generalisation ability. \newline

\section{Conclusion} \label{Conclusion}

In this paper, we presented a novel machine learning approach to predict patient treatment adherence using a large and reliable dataset comprising 165,223 historic injection disposal records collected from 5,915 HealthBeacon SSB units. The machine learning model construction methodology was discussed in detail and the proposed approach was validated by generating and evaluating various performance metrics on a real-world testing set. 

We demonstrate that HealthBeacon SSB data can be used to predict if a patient is likely to fail to take their medication on time with an accuracy of 81.3\%. The model demonstrated very good performance in predicting patient adherence with a ROC AUC score of 0.86 for a new patient dataset.

The results of this study validate that the data collected from the HealthBeacon SSB in combination with a machine learning model provides an accurate way of identifying patients who are at risk for future non-adherence. These valuable insights will enable targeted patient interventions.  \newline

\section{Ethics Statement}
The study was classified as a service evaluation and optimization project using irrevocably anonymized data, which does not require ethical approval or consent.\newline
\section{Acknowledgement}

This work was completed as part of the “Predictive analytics for patient treatment adherence” project funded by Enterprise Ireland and HealthBeacon Ltd. (IP 2018 0764), research carried out in association with the Insight SFI Research Centre for Data Analytics   (SFI/12/RC/2289\_P2). This project is co-funded by the European Regional Development Fund (ERDF) under Ireland’s European Structural and Investment Funds Programmes 2014-2020. The authors would also like to thank Dr. Amy Hall for proofreading and providing useful feedback on the manuscript.\newline

\bibliographystyle{IEEEtran}
\bibliography{refs}             

\end{document}